\documentclass[11pt]{llncs}
\usepackage[utf8]{inputenc}

\usepackage{authblk}
\usepackage{amssymb}
\usepackage{adjustbox}
\usepackage{float}
\restylefloat{table}
\usepackage{color,soul}
\usepackage{aliascnt}
\usepackage{amsmath}
\usepackage{mathtools}
\usepackage{textgreek}
\usepackage[a4paper, total={6.14in, 9.21in}]{geometry}
\usepackage{geometry}
\usepackage{ebgaramond,newtxmath}
\usepackage{glossaries}
\usepackage{chngcntr}
\counterwithin{figure}{section}
\usepackage{indentfirst}

\makeglossaries

\newacronym{AI}{AI}{Artificial Intelligence}
\newacronym{XAI}{XAI}{Explainable Artificial Intelligence}
\newacronym{DNN}{DNN}{Deep Neural Network}
\newacronym{VGG}{VGG}{Deep Neural Network}
\newacronym{LIME}{LIME}{Local Interpretable Model-agnostic Explanations}
\newacronym{ROAR}{ROAR}{RemOve And Retrain}
\newacronym{ReLU}{ReLU}{Rectified Linear Unit}
\newacronym{SHAP}{SHAP}{SHapley Additive exPlanations}
\newacronym{CAM}{CAM}{Class Activation Mapping}
\newacronym{GradCAM}{GradCAM}{Gradient-weighted Class Activation Mapping}
\newacronym{LRP}{LRP}{Layer Wise Relevance Propagation}

\newaliascnt{eqfloat}{equation}
\newfloat{eqfloat}{h}{eqflts}
\floatname{eqfloat}{Equation}
\newcommand*{\ORGeqfloat}{}
\let\ORGeqfloat\eqfloat
\def\eqfloat{%
  \let\ORIGINALcaption\caption
  \def\caption{%
    \addtocounter{equation}{-1}%
    \ORIGINALcaption
  }%
  \ORGeqfloat
}

\makeatletter
\newcommand{\printfnsymbol}[1]{%
  \textsuperscript{\@fnsymbol{#1}}%
}
\makeatother

\title{A Review of Explainable Artificial Intelligence in Manufacturing}

\author{Georgios Sofianidis\thanks{equal contribution}\inst{1} \and Jo\v{z}e M. Ro\v{z}anec\printfnsymbol{1}\inst{2,3} \and Dunja Mladeni\'{c}\inst{2} \and Dimosthenis Kyriazis\inst{1} }

\institute{Department of Digital Systems, University of Piraeus, Piraeus, Greece \\
\email{\{george.sofianidis,dimos\}@unipi.gr}
\and
Jo\v{z}ef Stefan Institute, Jamova 39, 1000 Ljubljana, Slovenia\\
\email{\{joze.rozanec,dunja.mladenic\}@ijs.si}
\and
Jo\v{z}ef Stefan International Postgraduate School, Jamova 39, 1000 Ljubljana, Slovenia
}

\date{\today}

\begin{document}

\maketitle

\thispagestyle{plain}\pagestyle{plain}

\begin{abstract}
The implementation of Artificial Intelligence (\acrshort{AI}) systems in the manufacturing domain enable higher production efficiency, outstanding performance, and safer operations, leveraging powerful tools such as deep learning and reinforcement learning techniques. Despite the high accuracy of these models, they are mostly considered black boxes: they are unintelligible to the human. Opaqueness affects trust in the system, a factor that is critical in the context of decision-making. We present an overview of Explainable Artificial Intelligence (\acrshort{XAI}) techniques as a means of boosting the transparency of models. We analyze different metrics to evaluate these techniques and describe several application scenarios in the manufacturing domain.
\end{abstract}


\section{Introduction}

The increasing digitalization of every aspect of life provides vast amounts of data, enabling the implementation of Artificial Intelligence (\acrshort{AI}) models. The manufacturing and process industry is not an exception to this trend. \acrshort{AI} models play a significant role in many aspects of the manufacturing process. \acrshort{AI} models drive better quality by enhancing quality inspection and process monitoring in production lines, ease reconfiguration and customization of automated part handling, fault diagnosis and event prediction, more agile production management, flexible production planning, and enabling safe collaboration between humans and cobots. Especially the latter is a big step towards the transition into Industry 5.0, where the focus is on the synergy between humans and robots and the actors are collaborators instead of competitors.

\acrshort{AI} models provide the means to automate many tasks and achieve unprecedented performance levels. However, in most cases, such models are opaque to the user: they work as black-boxes. Their predictions are mostly accurate, but no intuition behind the reasoning process is available to human users. Given the impact of those predictions on the decision-making processes, it is crucial to develop mechanisms and techniques to provide insights to users on such an \acrshort{AI} model reasoning process. The development of such techniques and mechanisms and how those insights are presented has given birth to a research field of its own, known as Explainable Artificial Intelligence (\acrshort{XAI}). While the field of \acrshort{XAI} can be traced back to the 1970's \cite{scott1977explanation}, it has experienced a new flourishment since the rise of modern deep learning\cite{xu2019explainable}. 

Though there is no single definition of the scope of this research field, most authors agree it includes intrinsically interpretable models and post-hoc explainability models (the model's capability of being explained by another interpretable model). Authors identify two sources of model opacity (or opaqueness)\cite{chan2021explainable}: (i) the complexity of the formal structure of the model is beyond human comprehension, or alien to human reasoning, or (ii) because the inner workings of the model cannot be shared (e.g., being considered a trade secret). Model opaqueness can be relative to expert knowledge: e.g., it can be opaque to an analyst but not to the machine learning engineer. \cite{muller2021deep} introduced the term \textit{deep opacity} to describe models whose opacity cannot be removed even by human experts.
When presenting insights on the reasoning process of an \acrshort{AI} model, the explanations should resemble a logic explanation\cite{samek2019towards}, and take into account relevant context. \cite{henin2021multi} considers context has three elements related to the explainee: (i) \textit{Profile} (user profile, to whom we present the explanation), (ii) \textit{Objective} (refer to the goals of the explanation, e.g., are the explanations meant to improve the model, enhance trust in the system, aid on decision-making or foster action based on decisions made), and (iii) \textit{focus} (if the explanation is either global or local). In local explanations, the specific point of interest must be considered part of the context. When the explanations aim to aid decision-making or take action, they should provide information regarding actionable features.

\begin{figure}[h]
    \centering
    \includegraphics[width=0.95\textwidth]{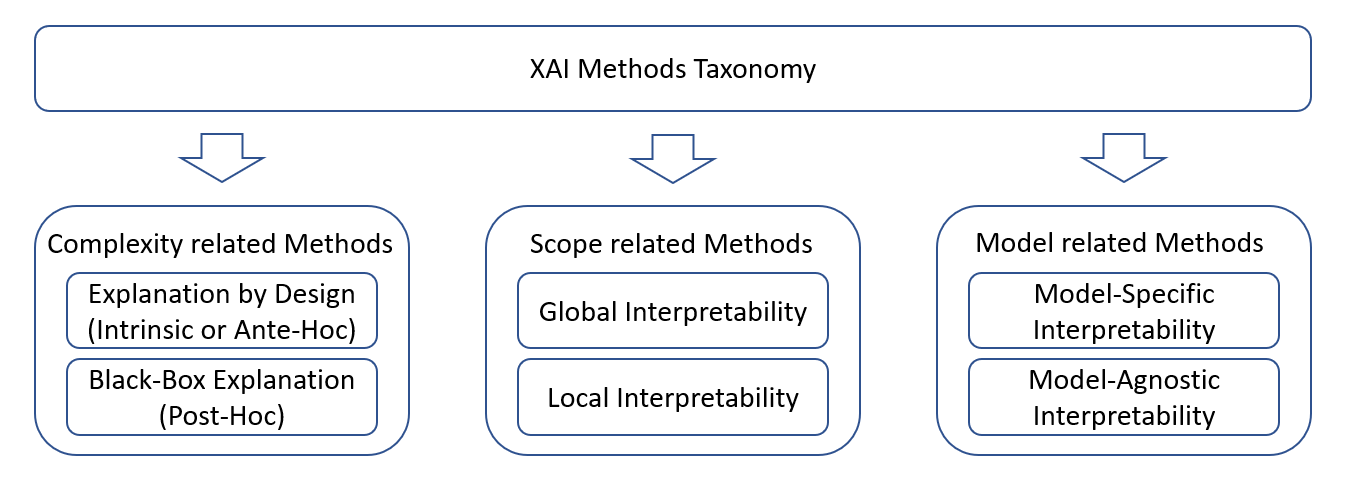}
    \caption{XAI taxonomy}
    \label{fig:XAI-taxonomy}
\end{figure}

\acrshort{XAI} techniques and methods can be classified into three categories, considering the explainability source, the scope of the explanation, and the level of dependency on the forecasting model used (see fig. \ref{fig:XAI-taxonomy}). We distinguish intrinsically explainable models and forecasting models that require post-hoc models to get insights into the forecast's reasoning process regarding the explainability source. Concerning the explanation's scope, explanations can be global (describe the behavior of the whole model for the average of forecasts provided) or local (describe the model's behavior for a particular forecast). Finally, regarding the dependency on the forecasting model's explanation, we distinguish model-agnostic (can be applied to any \acrshort{AI} model) or model-specific techniques (can be applied only to \acrshort{AI} models built with a particular algorithm or type of algorithms).

In this chapter, we introduce the field of Explainable Artificial Intelligence, describing methods and techniques used to identify meaningful features driving forecasts, current approaches used to evaluate such models, applications and use cases in the industrial domain, and open challenges. When doing so, we do not consider intrinsically explainable models. 

\section{Methods and techniques}

Different methods and techniques have been introduced to boost the transparency and acceptance of \acrshort{AI} models and different taxonomies have been proposed in literature based on the explanation generating mechanism, the type of explanation, the scope of explanation, the type of model it can explain, or a combination of these features. \cite{arrieta2020explainable} classified those methods into intrinsic interpretable models and post-hoc explanations and divided the latter to text explanations, visual explanations, local explanations, explanations by example, explanations by simplification, and feature relevance explanations techniques. 
\cite{bodria2021benchmarking} introduced a categorization of explanation methods based on the type of explanation returned and divided them based on the most common data types such as tabular, image, and text. For tabular data, feature importance is one of the most popular types of explanation returned by local explanation methods. The explainer assigns to each feature an importance value which represents how much that particular feature was important for the prediction under analysis. The sign and magnitude of each importance value are also considered to understand the contribution of each feature. Similar to the above but in the field of image classification, saliency maps can be used as explanations. Those are modeled as matrices with the same dimensions as that of the image we want to explain, and each element of the matrix represents the saliency of each pixel to the forecast. Another type of explanation that can be implemented on tabular data is the rule-based explanation. Human readable decision rules can give the end-user an explanation about the reasons that lead to the final prediction. A decision or factual or logic rule is a set of premises that lead to a specific forecast. Counterfactual rules are a set of rules that lead to the opposite of a specific forecast.
\cite{markus2020role} classified \acrshort{XAI} techniques according to the type of explanation and the scope of explanation. The three types he distinguished are model-based, attribution-based, and example-based explanations. In this chapter, we present some of the well-known explainability methods based on the taxonomy introduced by \cite{markus2020role}.

The class of \textit{model-based explanations} include methods that are either explainable by nature (intrinsic explainability) or methods that use a different interpretable model to explain the task model (post-hoc explainability). The first subclass can be divided into sparse linear classifiers (e.g., linear or logistic regression, generalized additive models (GAMs)), discretization methods (e.g., rule-based learners, decision trees), and example-based models (e.g., K-nearest neighbors). The second subclass includes interpretable surrogate models that can approximate the task model and can be used as post-hoc explanations.

The class of \textit{attribution-based explanations} use the explanatory power of input features to explain the task model. These approaches are also known as feature (a.k.a variable) importance, relevance, or influence methods. Most post-hoc explanations fall under this category which can further be divided into perturbation-based and backpropagation-based methods. 

Among the perturbation-based methods, we can find the \textit{Prediction Difference Analysis (PDA)} \cite{robnik2008explaining}, which is based on the idea that the relevance of an input feature concerning the class can be estimated by measuring how the predictions change if this particular feature is removed. This method cannot deal with saturated classifiers (models whose output does not change after removing part of the features). A similar approach for images was developed by \cite{zeiler2014visualizing} with the  \textit{Deconvolutional Networks}, which attempts to reconstruct the feature map into the layer input or the original image. The proposed networks used convolution, max-pooling layers, and the \acrshort{ReLU} activation function. Sliding a gray-color square over the image, they measure changes in feature activations and the classification scores. 
A variation of this method was developed by \cite{fong2017interpretable}, who, instead of using a gray-square, replaces regions of an image with constant values, noise, or performs some blurring on the image. This method was evolved by \cite{petsiuk2018rise}, who chose upsampled, random binary masks to perform the occlusions and analyzed their impact on the target class classification score. Another variation of \cite{zeiler2014visualizing} was introduced by \cite{zintgraf2017visualizing}, who removed several features at once by using prior knowledge about images and choosing patches of connected pixels as feature sets to analyze the effects of different window sizes on top scoring classes. The huge computational cost of this method was later minimized by \cite{gu2019contextual} through the  \textit{Contextual Prediction Difference Analysis}, which also solved the problem of saturated classifiers by producing a model-aware saliency map.

Another family of explainability methods computes feature attributions from a forward or backward pass through the network. They require architectural or backpropagation rule modifications or access to intermediate layers. However, most of these methods have lower computational costs than the ones mentioned above, leading to faster results. One of the first approaches of this kind was introduced by \cite{simonyan2013deep}, who computed feature attributions by taking the partial derivative of the output class with respect to the input. The resulting absolute values allow identifying which input features can be perturbed the least for the output to change the most. A drawback of this method is that it is noisy, and the absolute value of the gradients prevents the detection of positive and negative evidence in the input. This approach was improved by the \textit{Gradient * Input} method\cite{shrikumar2016not}, which increases the sharpness of attribution maps by taking the signed partial derivatives of the output with respect to the input and multiplying feature-wise by the input itself. The multiplication with the input indicates the interest in the salience rather than sensitivity. \cite{shrikumar2016not} introduced the \textit{Deep Learning Important FeaTures (DeepLIFT)} method, which uses a derivative-based method to propagate activation differences instead of gradients through the network. The intuition behind the method is that though the partial derivatives do not explain a single decision, they indicate what change in the image could make a change in the prediction. In the same line, \cite{sundararajan2017axiomatic} developed the \textit{Integrated Gradients} approach, which relies on the idea of computing attributions by multiplying the input variable element-wise with the average partial derivative, as the input varies from a baseline to its final value. \textit{Smooth-Grad}\cite{smilkov2017smoothgrad}takes a different approach, and focuses on local sensitivity, and calculates averaging maps with a smoothing effect made from several small perturbations of an input image. The effect is enhanced by further training with these noisy images. Finally, it sharpens the sensitivity maps, to increase their quality. \cite{zeiler2014visualizing} was evolved by \cite{springenberg2014striving}, who proposed the \textit{All Convolutional Net}, as an alternative that replaces the max-pooling layer for convolutional layers with an increased stride. A slightly different approach was proposed by \cite{zhou2016learning}, who introduced the \textit{Class Activation Mapping (\acrshort{CAM})}. This method relies on the observation that some convolutional layers behave as unsupervised object detectors, and it uses global average pooling to create heat maps of a pre-softmax layer. The heat maps point out the regions of an image that are responsible for a prediction. \textit{Gradient-weighted Class Activation Mapping (\acrshort{GradCAM})}\cite{selvaraju2017grad} uses the gradient information to understand how strongly does each neuron activate in the last convolutional layer of the neural network. The localizations are combined with existing high-resolution visualizations to obtain high-resolution class-discriminative guided visualizations as saliency masks. The \acrshort{CAM} and \acrshort{GradCAM} approaches inspired the \textit{GradCAM++} method\cite{chattopadhay2018grad}, which combines the positive partial derivatives of feature maps of a rear convolutional layer with a weighted special class score to explain the occurrence of multiple object instances in an image. \textit{Layer Wise Relevance Propagation (\acrshort{LRP})} \cite{bach2015pixel} is a gradient method suffering from vanishing gradient problems. The main idea behind this is the decomposition of the prediction function as a sum of layer-wise relevance values. The prediction is redistributed backward using local redistribution rules until assigning a relevance score to each input feature. There are different variations of the \acrshort{LRP} algorithm based on the backward redistribution rule.

Many explainability methods were built, relying on surrogate models to provide explanations regarding the reference model. One of such methods is \textit{TREPAN} \cite{craven1995extracting} which provides heuristics to issue queries against neural networks and create a decision tree that approximates forecasts from the given network, while providing an interpretable set of rules that explain the forecast. A more general approach was presented in the \textit{Local Interpretable Model-agnostic Explanations (\acrshort{LIME})}\cite{ribeiro2016should}, which can explain the predictions of any \acrshort{AI} model through a post-hoc, local, linear, and interpretable model. The model attempts to learn a particular forecast, by matching the given feature vector and perturbed inputs, to the results obtained from the reference model. Since the creation of \acrshort{LIME}, multiple variants were developed. \textit{k-LIME} (\cite{hall2017machine}) uses local generalized linear model surrogates to explain the predictions, while local regions are defined by k clusters instead of perturbed samples. The criteria to define the value of k is to K is that predictions from the local generalized linear models maximize R\textsuperscript{2}. In addition to this, a global surrogate linear generalized model is trained to provide information about overall feature average trends. \textit{DLIME} (\cite{zafar2019dlime}) proposes a deterministic version of \acrshort{LIME}, where instead of random perturbations, they apply agglomerative hierarchical clustering to group the training data. The hierarchical clustering does not require prior knowledge regarding clusters. A dendrogram is cut where the gap is the largest between two successive groups to determine the number of clusters. A k-Nearest Neighbour classifier is trained to classify new instances into those clusters based on the clusters obtained. All data points belonging to a given cluster are used to train a linear model, which provides deterministic and consistent local explanations. \textit{LIMEtree} (\cite{sokol2020limetree}) follows a similar approach to \acrshort{LIME}, building a regression tree as surrogate model. The regression tree enables capturing non-linear relationships between the interpretable features and the target variable. At the same time, it does not require independence between interpretable features. The authors consider the model's biggest advantage is providing personalized counterfactual explanations through an interactive interface that enables imposing certain conditions on the sample of interest. Inspired in LIME, \cite{elenberg2017streaming} developed STREAK, an interpretability method for neural networks conceived as a set function maximization, achieving similar accuracy than LIME, while having a faster runtime execution. A slightly different approach is presented in Anchors\cite{ribeiro2018anchors}, where a set of rules replaces the surrogate model. Since the local behavior of a model can be highly non-linear, the authors propose using a set of if-then rules, which are intuitive and easy to understand. To explore the model's behavior in the perturbation space, the authors apply multi-armed bandits to incrementally construct the rules, generate candidate predicates, and choose the one with the highest precision until a given precision threshold is reached with a high probability. \textit{LoRE - Local Rule-Based Explanations}\cite{guidotti2018local} proposes a parameter-free, two step method that also provides rule-based explanations. First, it creates a balanced set of neighbor instances using a genetic algorithm to explore the decision boundary of the data point of interest. Then it builds a decision tree classifier, which enables to derive decision rules and counterfactuals. \textit{Local Foil Trees}\cite{van2018contrastive} specifically deal with generating counterfactual explanations. To that end, they consider two possible outputs: the model forecast (fact), and the desired label (foil). A decision tree is then built based on the local dataset. The rules are computed from the difference between paths regarding the \textit{"fact leaf"}, and \textit{"foil leaf"}.

While most explainability methods based on surrogate models provide specific techniques, \cite{henin2019towards} developed a framework that enabled comparing surrogate models on three dimensions: data sampling, explanation generation, and interaction. \cite{sokol2019blimey} considered a slightly different approach and developed an algorithmic framework (\textit{bLIMEy - build \acrshort{LIME} yourself}) that enables building custom local surrogate explainers for model predictions, considering three dimensions: data sampling, explanation generation,  and interpretable representation.

Another local-agnostic explanation method is \textit{\acrshort{SHAP}} \cite{lundberg2017unified} which stands for SHapley Additive exPlanations and can be used to produce several explanation models. These models compute \acrshort{SHAP} values: a unified measure of feature importance based on the Shapley values, a concept from cooperative game theory. The different explanation models proposed by \acrshort{SHAP} differ on how they approximate the computation of the \acrshort{SHAP} values. The explanation models provided by \acrshort{SHAP} are called \textit{additive feature attribution methods}. The construction of the \acrshort{SHAP} values allows to employ them both locally, in which each observation gets its own set of \acrshort{SHAP} values, and globally, by exploiting collective \acrshort{SHAP} values.

In the image classification field, two explanators can be implemented for deep networks: DEEP-SHAP and GRAD-SHAP. DEEP-SHAP is a high-speed approximation algorithm for shap values in deep learning models that connect with the DeepLift algorithm. The implementation is different from the original DeepLift by using a baseline distribution of background samples instead of a single value and using Shapley equations to linearise non-linear components of the black-box such as max, softmax, products, divisions. GRAD-SHAP, instead, is based on IntGrad and SmoothGrad algorithms. IntGrad values are a bit different from SHAP
values, and require a single reference value to integrate from. As an adaptation to approximate SHAP values, GRAD-SHAP reformulates the integral as an expectation and combines that expectation with sampling reference values from the background dataset as done in SmoothGrad.

Another family of explainability techniques is that of \textit{example-based explanations}. Methods in this class explain the task model by selecting particular instances from the dataset that describe the model or by creating new instances. Instances that are well predicted by the forecasting model (prototypes) and instances that are not well predicted by the model (criticism) are the influential instances for the model parameters or output, while counterfactual explanations indicate the required changes in the input side that will have significant changes (e.g., reverse the prediction) in the prediction/output. \cite{kim2016examples} proposed a methodology named \textit{MMD-CRITIC} to learn prototypes and criticisms for a given dataset using the maximum mean discrepancy (MMD) as a measure of similarity. \cite{plumb2018model} introduced \textit{MAPLE}. This post-hoc local agnostic explanation method can also be used as a transparent model due to its internal structure. It combines random forests with feature selection methods to return feature importance-based explanations. \textit{DICE} which stands for Diverse Counterfactual Explanations \cite{mothilal2020explaining} is a local, post-hoc and agnostic method that solves an optimization problem with several constraints to ensure feasibility and diversity when returning counterfactuals. Feasibility is critical in the context of counterfactuals since it allows avoiding examples that are unfeasible.

We classify the aforementioned methods according to multiple criteria in Table \ref{T:XAI-CLASSIFICATION}.

\begin{table*}[t]
\centering
\resizebox{\columnwidth}{!}{
\begin{tabular}{|l|c|l|c|l|c|c|c|}
\hline
\textbf{Explanation technique} & \textbf{Reference} & \multicolumn{1}{c|}{\textbf{Model based}} & \textbf{Attribution based} & \multicolumn{1}{c|}{\textbf{Example based}} & \textbf{\begin{tabular}[c]{@{}c@{}}Local (L) /\\ Global (G)\end{tabular}} & \textbf{\begin{tabular}[c]{@{}c@{}}Agnostic (A) /\\ Specific (S)\end{tabular}} & \textbf{Data Type} \\ \hline
All Convolutional Net & \cite{springenberg2014striving} & \multicolumn{1}{c|}{X} & X &  & L & S & IMAGE \\ \hline
Anchors & \cite{ribeiro2018anchors} &  & X &  & L/G & A & TABULAR/TEXT \\ \hline
Class Activation Mapping (\acrshort{CAM}) & \cite{zhou2016learning} &  & X &  & L & S & IMAGE \\ \hline
Contextual Prediction Difference Analysis & \cite{fong2017interpretable} &  & X &  & L & S & IMAGE \\ \hline
Deconvolutional Networks & \cite{zeiler2014visualizing} & \multicolumn{1}{c|}{X} & X &  & L & S & IMAGE \\ \hline
Deep Learning Important FeaTures (DeepLIFT) & \cite{shrikumar2016not} &  & X &  & L & S & ANY \\ \hline
DICE & \cite{mothilal2020explaining} &  & \multicolumn{1}{l|}{} & \multicolumn{1}{c|}{X} & L & A & ANY \\ \hline
DLIME & \cite{zafar2019dlime} & \multicolumn{1}{c|}{X} & X &  & L & A & ANY \\ \hline
GradCAM++ & \cite{chattopadhay2018grad} &  & X &  & L & S & IMAGE \\ \hline
Gradient & \cite{simonyan2013deep} &  & X &  & L & S & ANY \\ \hline
Gradient * Input & \cite{shrikumar2016not} &  & X &  & L & S & ANY \\ \hline
Gradient Weighted Class Activation Mapping (\acrshort{GradCAM}) & \cite{selvaraju2017grad} &  & X &  & L & S & IMAGE \\ \hline
Integrated Gradients & \cite{sundararajan2017axiomatic} &  & X &  & L & S & ANY \\ \hline
k-LIME & \cite{hall2017machine} & \multicolumn{1}{c|}{X} & X &  & L & A & ANY \\ \hline
Layer Wise Relevance Propagation (\acrshort{LRP}) & \cite{bach2015pixel} &  & X &  & L & A & ANY \\ \hline
LIME & \cite{ribeiro2016should} & \multicolumn{1}{c|}{X} & X &  & L & A & ANY \\ \hline
LIMETree & \cite{sokol2020limetree} & \multicolumn{1}{c|}{X} & X &  & L & A & TAB \\ \hline
Local Foil Trees & \cite{van2018contrastive} & \multicolumn{1}{c|}{X} &  & \multicolumn{1}{c|}{X} & L & A & TABULAR \\ \hline
LoRE & \cite{guidotti2018local} &  & X &  & L & A & TABULAR \\ \hline
MAPLE & \cite{plumb2018model} & \multicolumn{1}{c|}{X} & X & \multicolumn{1}{c|}{} & L & A & TABULAR \\ \hline
Meaningfull Perturbation & \cite{fong2017interpretable} &  & X &  & L & S & IMAGE \\ \hline
MMD-CRITIC & \cite{kim2016examples} &  & \multicolumn{1}{l|}{} & \multicolumn{1}{c|}{X} & G & A & ANY \\ \hline
Prediction Difference Analysis (PDA) & \cite{robnik2008explaining,zintgraf2017visualizing} &  & X &  & L & S & IMAGE \\ \hline
RISE & \cite{petsiuk2018rise} &  & X &  & L & S & IMAGE \\ \hline
SHAP & \cite{lundberg2017unified} &  & X &  & L/G & A & ANY \\ \hline
Smooth Grad & \cite{smilkov2017smoothgrad} &  & X &  & L & S & IMAGE \\ \hline
STREAK & \cite{elenberg2017streaming} &  & \multicolumn{1}{l|}{} &  & L & A & IMAGE \\ \hline
TREPAN & \cite{craven1995extracting} &  & X &  & G & S & TABULAR \\ \hline
\end{tabular}}
\caption{Classification of \acrshort{XAI} techniques. \label{T:XAI-CLASSIFICATION}}
\end{table*}

\section{Evaluation Measures}

Explainability is considered a subjective concept. \cite{markus2020role} considers that an \acrshort{AI} system is explainable if either the model is intrinsically interpretable or if the non-interpretable model can be complemented with an interpretable and faithful explanation. While the \acrshort{XAI} techniques provide different kinds of information, the perceived quality of the explanations depends on the users, the domain, the information of interest, and the explanation itself. To evaluate the explanations, it is necessary to define different criteria of goodness for an explanation. Given an interpretable approximation for a reference, model \cite{lakkaraju2017interpretable} lists four aspects to be considered on evaluation: fidelity (ability to capture the reference model behavior correctly), unambiguity (ability to provide a single and deterministic rationale to explain each data instance), interpretability (the approximation should be human-understandable), and interactivity. The aspect of fidelity is further elaborated by \cite{kulesza2013too}, who considers two properties: soundness (the extent to which each explanation component is truthful to the reference model) and completeness (the extent to which the explanation describes the reference model). \cite{ylikoski2010dissecting} enumerate another three criteria: sensitivity, the degree of integration, and cognitive salience. Sensitivity is defined as the strength of the relationship of explanatory variables with background conditions: the weaker the relationship, the more convincing the explanation. The degree of integration refers to the connectedness of the explanation to a larger theoretical framework. Finally, cognitive salience is defined as the ease with which the rationale behind the explanation can be followed.

The aforementioned criteria require different evaluation approaches. \cite{doshi2017towards} identified three categories of them: 
\begin{itemize}
    \item \textbf{Application-grounded evaluation}: grounded in a real-world application, collects domain expert's feedback regarding the explanations provided to them. 
    \item \textbf{Human-grounded evaluation}: refers to feedback obtained from experiments performed with lay users, when no real-world application exists in place.
    \item \textbf{Functionality-grounded evaluation}: the evaluation is performed considering some formal definition or criteria, that measures the explanation quality. 
\end{itemize}

To assess the explainability methods, \cite{guo2018lemna} propose three tests for functionality-grounded evaluations: \textbf{Feature Augmentation Test}, \textbf{Synthetic Test}, and \textbf{Feature Deduction Test}. The \textbf{Feature Augmentation Test} considers that if the values of the explainable features from a specific instance are replaced by the values of those features from an instance with a different label (e.g., "new-label"), the classification outcome should be  "new-label". The \textbf{Synthetic Test} is based on the assumption that if the explainability features are accurately selected, new synthetic instances can be created by preserving the explainability feature values and assigning random values to the rest of the features without affecting the forecast outcome. Finally, the \textbf{Feature Deduction Test} considers that if the selected explainability features are correctly selected, removing one of them from the input should lead to a different forecast. Even though this approach is frequently adopted in the literature\cite{zeiler2014visualizing,fong2017interpretable,zintgraf2017visualizing,petsiuk2018rise}, \cite{hooker2018benchmark} pointed out that samples, where a subset of features are removed have a different data distribution than the samples the model was trained on, violating a key machine learning assumption. They instead propose the RemOve And Retrain (\acrshort{ROAR}) approach, which for each feature deemed important, they replace it by a non-informative value in the train and test sets, retrain the model and measure the performance change. In addition to this technique, they propose using a random assignment of feature importance as a benchmark to measure the quality of explainability feature extraction techniques.

There is currently little research regarding application and human-grounded evaluations\cite{doshi2017towards,zhou2021evaluating}. A popular and domain-specific method is to evaluate to create a heatmap regarding model sensitivity to region-based perturbations. According to the heatmap, the main idea behind this is that the perturbation of relevant input variables would lead to a decline in prediction score than the perturbation of input features with less importance.  \cite{kulesza2013too} used questionnaires with short responses and Likert scales. In contrast, \cite{lage2019evaluation} used three quantitative metrics: accuracy, response time, and subjective satisfaction. The authors measured accuracy and response time regarding the subject response to different tasks proposed in their research. Subjective satisfaction was measured on a Likert scale for each explanation. \cite{lage2018human} proposed the Human Interpretability Score (HIS - see Eq. \ref{E:HIS}), which constitutes an alternative metric regarding the user's response time. On the other side, there is a wider set of metrics reported for functionality-grounded evaluations.

\begin{eqfloat}[H]
\begin{equation}\label{E:HIS}
    HIS(x, R)= \begin{dcases}
    0,& \text{if } RT_{mean}(x, R) > RT_{max}\\
    RT_{max} - RT_{mean}(x, M), & RT_{mean}(x, R) \leq RT_{max}
\end{dcases}
\end{equation}
\caption{Human Interpretability Score. Measures how long it takes the user to predict the label
assigned to certain data point, assigning a cap to the response time. \textit{x} and \textit{R} correspond to the instance and model considered.}
\end{eqfloat}

Among the metrics proposed by \cite{nguyen2020quantitative} we find \textit{Mutual Information}, \textit{Diversity}, \textit{Monotonicity}, \textit{Non-sensitivity}, and \textit{Effective complexity}. 
\textit{Mutual Information} is considered when creating an interpretable data representation. \cite{nguyen2020quantitative} proposes measuring Mutual Information on two cases: (i) between the features of the original model and the subset of explainable features, and (ii) against the target values. Ideally, the number of explainable features should be reduced to maximize simplicity and broadness, while aiming towards keeping a high fidelity regarding the target label (see Eq. \ref{E:MUTUAL-INFORMATION}).

\begin{eqfloat}[H]
\begin{equation}\label{E:MUTUAL-INFORMATION}
    I(x, y)= D_{KL} (P_{(x,y)} \parallel P_{x} \otimes P_{y})
\end{equation}
\caption{Mutual Information. Measures the mutual dependence between two random variables \textit{x} and \textit{y}.}
\end{eqfloat}

\textit{Diversity} attempts to measure the degree to which a set of rules integrates to the explanation (see Eq. \ref{E:DIVERSITY}).
\textbf{Monotonicity} considers that feature attributions should be monotonic. \cite{nguyen2020quantitative} proposes measuring it as the Spearman's correlation between two vectors: (i) the absolute values of attributions, and (ii) the corresponding expectations.
The intuition behind the \textbf{Non-sensitivity} metric (see Eq. \ref{E:NON-SENSITIVITY}) is to assess that the explainability method does not assign any relevance score to the features the model is not functionally dependent on. The authors compute it as the cardinality of the symmetric difference between features assigned zero attribution and the features the model does not functionally depend on.
\textbf{Effective complexity} measures if some explanation features can be ignored without significantly affecting the prediction (see Eq. \ref{E:EFFECTIVE-COMPLEXITY}).

\begin{eqfloat}[H]
\begin{equation}\label{E:DIVERSITY}
    Diversity = \sum_{x_i, x_j \in E; x_i \neq x_j} \frac{d(x_i,x_j)}{2N\textsubscript{E}} 
\end{equation}
\caption{Diversity metric. \textit{E} is the set of examples considered, \textit{d} is a distance metric for the space \textit{X}, while \textit{N\textsubscript{E}} corresponds to the number of examples.}
\end{eqfloat}

\begin{eqfloat}[H]
\begin{equation}\label{E:NON-SENSITIVITY}
    |A_{0} \bigtriangleup X_{0}|
\end{equation}
\caption{Non-sensitivity. $A_{0}$ represents featues with zero attribution, $X_{0}$ refers to features on which the model is not functionally dependent on. $|\cdot|$ denotes the set cardinality, and $\bigtriangleup$ the symmetric set difference.}
\end{eqfloat}

\begin{eqfloat}[H]
\begin{equation}\label{E:EFFECTIVE-COMPLEXITY}
    k* = argmin_{k \in 1, ..., N} |M_{k}|\; where\; E(l(y*, f-M_{k})|x*_{M_{k}}) < \varepsilon
\end{equation}
\caption{Effective Complexity. $M_{k}$ denotes the set of top \textit{k} features, \textit{x} denotes features, $\varepsilon > 0$ corresponds to some arbitrary tolerance, $f-M_{k}$ is the restriction of the model \textit{R} to non-important features, given $M_{k}$.}
\end{eqfloat}

The \textbf{Local Approximation Accuracy} was proposed by \cite{guo2018lemna} to compare the decision boundary of the surrogate model against the original one. The authors do so by computing the Root Mean Squared Error between the original and surrogate model predictions on the test samples. A similar intuition is present in the \textbf{Disagreement} metric proposed by \cite{lakkaraju2017interpretable}. For a classification setting, they attempt to measure the surrogate model fidelity by computing the disagreement between labels of the surrogate model and the original one (see Eq. \ref{E:DISAGREEMENT}).

\begin{eqfloat}[H]
\begin{equation}\label{E:DISAGREEMENT}
    Disagreement(R) = \sum_{i=1}^{N} \Big| {x \big| x\in D, x satisfies\; q_i \wedge s_i, B(x) \neq c_i} \Big| 
\end{equation}
\caption{Disagreement metric. Quantifies the disagreement between a surrogate model R and the reference forecasting model \textit{B}, given a dataset \textit{D}. The triplet \textit{(q, s, c)} stands for (feature, operator, class).}
\end{eqfloat}

\cite{lakkaraju2017interpretable} propose another six metrics to evaluate forecast explanations: rule overlap, cover, the rule set size (see Eq. \ref{E:RULE-SET-SIZE}), the rule set maximum width, the number of descriptor sets, and feature overlap. The \textbf{Rule overlap} computes the overlap between pairs of rules defined in the surrogate model. It is expected that the lower the overlap, the lower the surrogate model ambiguity (see Eq. \ref{E:RULE-OVERLAP}). \textbf{Cover} is defined as the number of instances that match a given rule from the surrogate model (see Eq. \ref{E:COVER}). The \textbf{Maximum Width} refers to the maximum width obtained from computing the width over all the elements from the surrogate model. The authors define an element as either rule conditions or neighborhood descriptors (see Eq. \ref{E:MAX-WIDTH}). The authors define the \textbf{Number of Unique Descriptor Sets} as the number of unique neighborhood descriptors provided in the surrogate model (see Eq. \ref{E:NUMBER-DESCRIPTORS}). Finally, the \textbf{Feature overlap} measures the features overlap between every pair of unique neighborhood descriptor and rule (see Eq. \ref{E:FEATURE-OVERLAP}). 

\begin{eqfloat}[H]
\begin{equation}\label{E:RULE-SET-SIZE}
    RuleSetSize(R) = NumberOfRules(q, s, c)
\end{equation}
\caption{Rule set size. \textit{R} denotes the decision set. The triplet \textit{(q, s, c)} stands for (feature, operator, class). The triplets are contained in the decision set.}
\end{eqfloat}

\begin{eqfloat}[H]
\begin{equation}\label{E:RULE-OVERLAP}
    RuleOverlap(R) = \sum_{i=1}^{N} \sum_{j=1, j \neq i}^{N} overlap(q_{i} \wedge s_{i}, q_{j} \wedge s_{j})
\end{equation}
\caption{Rule overlap. \textit{R} denotes the decision set. The triplet \textit{(q, s, c)} stands for (feature, operator, value).}
\end{eqfloat}

\begin{eqfloat}[H]
\begin{equation}\label{E:COVER}
    cover(R) = \Big| {x \big| x\in D, x satisfies\; q_i \wedge s_i, where i \in {1 ... N}} \Big| 
\end{equation}
\caption{Cover. \textit{R} denotes the decision set. The triplet \textit{(q, s, c)} stands for (feature, operator, value). \textit{D} represents a dataset, and \textit{x} and instance in such dataset.}
\end{eqfloat}

\begin{eqfloat}[H]
\begin{equation}\label{E:MAX-WIDTH}
    MaximumWidth(R) = max(width(e)), e \in \bigcup_{i=1}^{N} (q_{i} \cup s_{i})
\end{equation}
\caption{Maximum Width. \textit{R} denotes the decision set. e represents elements, which can be ether rule conditions or neighborhood descriptors.}
\end{eqfloat}

\begin{eqfloat}[H]
\begin{equation}\label{E:NUMBER-DESCRIPTORS}
    NumberOfUniqueDescriptorSets(R) = |dset(R)|, where\;dset(R) = \bigcup_{i=1}^{N} (q_{i})
\end{equation}
\caption{Number of Unique Descriptor Sets. \textit{R} denotes the decision set, and \textit{q} denotes features.}
\end{eqfloat}

\begin{eqfloat}[H]
\begin{equation}\label{E:FEATURE-OVERLAP}
    FeatureOverlap(R) = \sum_{i=1}^{N} FeatureOverlap(q, s_{i})
\end{equation}
\caption{Feature Overlap. \textit{R} denotes the decision set, \textit{q} denotes features in descriptor sets, and \textit{s} denotes operators.}
\end{eqfloat}

A different set of metrics is considered by \cite{poursabzi2018manipulating}, who for tree-based models measured the mean path length, the mean number of distinct features in a path, the number of nodes, and the number of nonzero features. Finally, \cite{slack2019assessing} reported assessing explainability methods based on the total number of runtime operation counts performed by the model when computing the forecast for a given input.

\section{Applications, Use Cases and Open Issues}

Though multiple \acrshort{XAI} methods exist, they do not suffice by themselves to provide human-understandable explanations. They are built into frameworks and applications that provide a convenient interface and additional context to achieve that goal. One such framework is bLIMEy\cite{sokol2019blimey}, which decomposes surrogate models into three steps: interpretable data representation (transform data from the original to the interpretable domain), data sampling, and explanation generation. \cite{henin2020generic} follows a similar approach and describes the IBEX (Interactive Black-box EXplanation system) framework with two components: an explainer that produces explanations based on user's needs, and a sampling component, that selects appropriate inputs to create the explanation. \cite{arya2020ai} describes \textit{AI Explainability 360}, an extensible toolkit developed that provides contextual explainers based on the stage of the \acrshort{AI} model development pipeline, kind of model, and explanation requirements. \cite{panigutti2020doctor} explores the usage of domain knowledge encoded in an ontology improves the quality of the explanations. \cite{rovzanec2021semantic} explores the usage of semantic technologies to abstract relevant concepts encoded in the features, avoid exposing sensitive details regarding the forecasting model, and provide higher-level information to the users. The authors complement model explanations with information regarding real-world events reported in the media that likely influenced the variables of interest. \cite{rovzanec2021xaikg} developed an ontology to model user's feedback based on a given forecast and provided explanations. \cite{zajec2021towards} developed an intelligent assistant for manufacturing, which creates directive explanations for the users using heuristics and domain knowledge. The application tracks user's implicit and explicit feedback regarding local forecast explanations, enabling application-grounded evaluations.

The integration of explainability methods into applications enables providing relevant information regarding model forecasts to different stakeholders. For instance, data scientists and machine learning engineers require low-level data to monitor the \acrshort{AI} model behavior, identify corner cases, and work towards a more accurate and robust model. On the other side, employees and supervisors require high-level insights that convey reasons behind the model forecasts, can interactively explore different \textit{"what-if"} scenarios, and provide feedback regarding the explanations provided. We envision explainability methods can be useful in a wide range of manufacturing use cases, such as automatic defect detection (inform the user on the image regions influencing the decision), production planning (provide an insight on the cost of the opportunity given different scheduling decisions), or demand forecasting (provide insights why we expect demand will take place and which factors affect the quantity estimates).

Several explainability techniques have been implemented in the manufacturing domain and specifically the predictive quality management domain (Quality 4.0) to boost the transparency of \acrshort{AI} deployed models. \cite{goldman2021explaining} used \acrshort{XAI} techniques such as \acrshort{CAM} and Contrastive gradient-based saliency maps to explain black-box classifiers in the area of quality welds in ultrasonically welded battery tabs. They produced heatmaps where they visualized several color maps to gain insights into true positive versus false-positive predictions. \cite{lee2021explainable} implemented several \acrshort{XAI} methods to provide explanations for domain experts in the area of defect classification of thin-film-transistor liquid-crystal display panels. Techniques such as \acrshort{CAM}, \acrshort{LRP}, integrated gradients, guided backpropagation, and SmoothGrad were implemented and visualized on a VGG-16 classification model. Based on the visualized results, \acrshort{LRP} and guided backpropagation were selected as they produced well-distributed heatmaps. Moreover, by fitting the model into a decision tree and converting the prediction results into human interpretable text, the authors achieved the maximum level of explainability when they presented the results to domain experts for evaluation purposes. 
In the area of manufacturing cost estimation, \cite{yoo2020explainable} described a method based on visualization of the machining features of a 3D computer aided design model that are influencing the increase in manufacturing costs. For the proposed purpose, a 3D gradient-weighted class activation mapping as \acrshort{XAI} method was applied.

Cybersecurity in a transversal concern related to all smart manufacturing cases. XAI techniques were successfully applied in the cybersecurity domain, to support the exploration of model vulnerabilities \cite{lapuschkin2019unmasking,ma2019explaining}, and identify perturbed data samples\cite{fidel2020explainability}.

In the European Horizon 2020 project STAR (Safe and Trusted Human Centric Artificial Intelligence in Future Manufacturing Lines), XAI is used to provide insights on most relevant features to each forecast, explore model vulnerabilities and help identify potential data poisoning. While providing accurate explanations to forecasts provides the users additional elements for decision-making, the vulnerabilities assessment and early data poisoning identification ensures the system is secure, enhancing users trust in the system.

\section{Conclusion}
The new industrial revolution relies on \acrshort{AI} to enable higher production efficiency, and safer operations. \acrshort{XAI} techniques provide means to reduce black-box models opaqueness, and increase trust in the system. In this contribution, we introduce the field of \acrshort{XAI}. We list several taxonomies found in the literature alongside state-of-the-art methods and techniques to interpret \acrshort{AI} models. We also include metrics with different qualitative and quantitative characteristics as a means of evaluating the above methods. Finally, we list applications of \acrshort{XAI}, describe several use cases in the manufacturing domain, and open opportunities. 

\acrshort{XAI} requires a multi-disciplinary approach. Special consideration needs to be given to understand how domain experts and end-users operate. Users must be involved in the \acrshort{XAI} outcomes validation. The integration of \acrshort{XAI} into manufacturing processes will be paramount for the transition into the fifth industrial revolution.

\section*{Acknowledgements}
This work has been carried out in the H2020 STAR project, which has received funding from the European Union's Horizon 2020 research and innovation programme under grant agreement No. 956573.

\bibliographystyle{plain}
\bibliography{main}

\printglossaries

\end{document}